\documentclass{article}
\usepackage{spconf,amsmath,graphicx}
\usepackage{url}
\usepackage{multirow}
\usepackage{amssymb}
\usepackage{multirow, threeparttable, booktabs, makecell}
\usepackage{color}
\usepackage{graphicx}
\usepackage{enumitem}
\usepackage{array}
\usepackage{placeins}
\usepackage{afterpage}
\usepackage[numbers,sort&compress]{natbib}

\title{ED-CEC: Improving rare word recognition using ASR postprocessing based on error detection and context-aware error correction}

\name{Jiajun He $^{\star}$ , Zekun Yang $^{ \ddagger}$, Tomoki Toda $^{ \ddagger}$}
\address{$^{\star}$ Graduate School of Informatics, Nagoya University, Japan\\ $^{ \ddagger}$ Information Technology Center, Nagoya University, Japan}

\copyrightnotice{To appear in Proc. ASRU2023}
\begin{document}
\ninept
\maketitle
\vspace{-2mm}
\begin{abstract}
\vspace{-1mm}
Automatic speech recognition (ASR) systems often encounter difficulties in accurately recognizing rare words, leading to errors that can have a negative impact on downstream tasks such as keyword spotting, intent detection, and text summarization. To address this challenge, we present a novel ASR postprocessing method that focuses on improving the recognition of rare words through error detection and context-aware error correction. Our method optimizes the decoding process by targeting only the predicted error positions, minimizing unnecessary computations. Moreover, we leverage a rare word list to provide additional contextual knowledge, enabling the model to better correct rare words. Experimental results across five datasets demonstrate that our proposed method achieves significantly lower word error rates (WERs) than previous approaches while maintaining a reasonable inference speed. Furthermore, our approach exhibits promising robustness across different ASR systems.
\end{abstract}
\vspace{-1mm}
\vspace{-3mm}
\section{Introduction}
\label{sec:intro}
\vspace{-2mm}

Automatic speech recognition (ASR) technology has made considerable progress in recent years, enabling machines to transcribe speech with marked accuracy \cite{li2022recent,meng2023jeit}. However, even with state-of-the-art (SOTA) ASR systems, there remains a persistent challenge in accurately recognizing rare words, such as named entities, technical terms, and specific names \cite{huber2021instant}. These rare words are often misrecognized as similar-sounding words in the recognition lexicon, resulting in errors that significantly degrade the overall transcription quality \cite{bekal2021remember}. Such errors can have a substantial impact on downstream tasks such as video summarization \cite{sharma2021speech} and named entity recognition \cite{cohn2019audio, baril2022named}. Consequently, improving the recognition of rare words has become a crucial objective in enhancing ASR performance.


To tackle the challenge of rare word recognition in ASR, several techniques have been proposed. These techniques primarily involve incorporating contextual knowledge into the ASR system \cite{huber2021instant, pundak2018deep, han2022improving, liu2022internal} and integrating an additional language model (LM) into the decoding phase to bias recognition results towards contextual knowledge \cite{williams2018contextual, le2021deep, zhou2022language, chen2022factorized}. In these approaches, contextual knowledge is typically represented by a list of words or phrases, known as contextual items, that are likely to appear in a given context. Various resources, such as lecture video slides, meeting minutes, and a user's contact book, can be utilized to construct the rare word list \cite{miranda2013improving, akita2015language}. However, these aforementioned approaches have certain limitations. On one hand, the method of incorporating contextual knowledge into the ASR system can be computationally expensive during both training and inference, and it may require significant modifications to the original ASR models' structure \cite{pundak2018deep}. Moreover, this approach may not effectively handle a large rare word list. 
On the other hand, the method of integrating an additional LM into the decoding phase necessitates careful weight tuning in different scenarios.

To address the limitations of previous approaches, ASR error correction (AEC) has been proposed \cite{zhang2023patcorrect, dutta2022error, wang2022towards, DBLP:journals/corr/abs-2208-04641, lin2023multi}. 
An AEC model is designed to be an independent model that does not alter the structure of the original ASR model, ensuring no risk of performance degradation. This characteristic makes it highly convenient to apply the AEC model across various domains, as it can be easily integrated by replacing the existing AEC model without the need for retraining the original ASR model \cite{wang2022towards}. Wang et al. \cite{wang2022towards} integrated contextual knowledge into an error correction model through a context encoder, which corrects the ASR output from scratch. However, this process raised significant concerns regarding inference speed. Interestingly, it was observed that the majority of words were identical between the ASR output and the ground truth.
Hence, Yang et al. \cite{DBLP:journals/corr/abs-2208-04641} proposed the use of an operation predictor to constrain the decoding process, resulting in a notable improvement in inference speed while retaining the capability to correct certain errors. However, owing to the lack of contextual knowledge integration, this approach could not effectively correct rare words.

\begin{figure}[t]
  \centering
  \includegraphics[width=1\columnwidth]{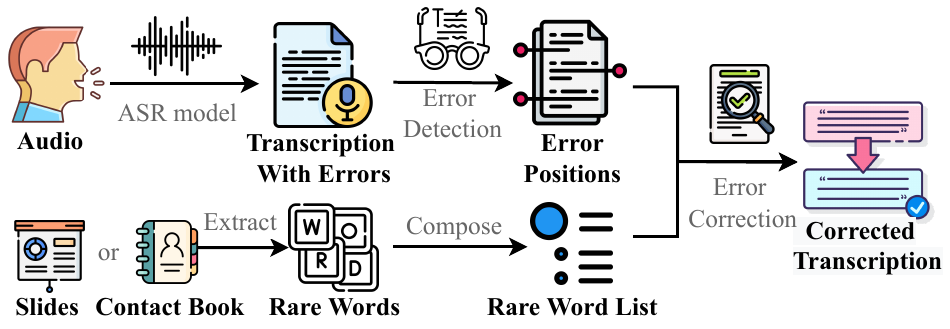}
  \vspace{-6mm}
  \caption{Pipeline of our proposed method.}
  \label{fig:pipeline}
\end{figure}

In this paper, we propose a novel method based on error detection and context-aware error correction (ED-CEC) to address the challenges associated with inference speed and rare word correction, as shown in Fig. \ref{fig:pipeline}. The rare word list used in the error correction module can be obtained from various sources such as slide texts or a contact book.
The contributions of this paper are summarized below:

\noindent
\begin{itemize}[leftmargin=*]
\vspace{-1.5mm}
\setlength{\topsep}{0pt}
\setlength{\itemsep}{0pt}
\setlength{\parsep}{0pt}
\setlength{\parskip}{0pt}
\item \noindent We propose a method to correct rare words based on ASR results. Our model includes an error detection module, which identifies incorrect positions and decodes only those positions to increase inference speed, and a context-aware error correction module, which corrects rare words by selecting relevant contextual items from a rare word list.
\item We conduct experiments on five datasets. The results demonstrate that our model achieves a relative word error rate reduction (WERR) ranging from 15.6$\%$ to 38.17$\%$ compared with the original ASR output and the average relative improvement in biased word error rate (B-WER) is 46.68\%. In addition, the proposed method achieves an inference speed of 2.8 $-$ 6.0 times higher than the previous SOTA model.
\vspace{-4mm}
\end{itemize}

\begin{figure*}[htbp]
  \centering
  \includegraphics[scale=0.133]{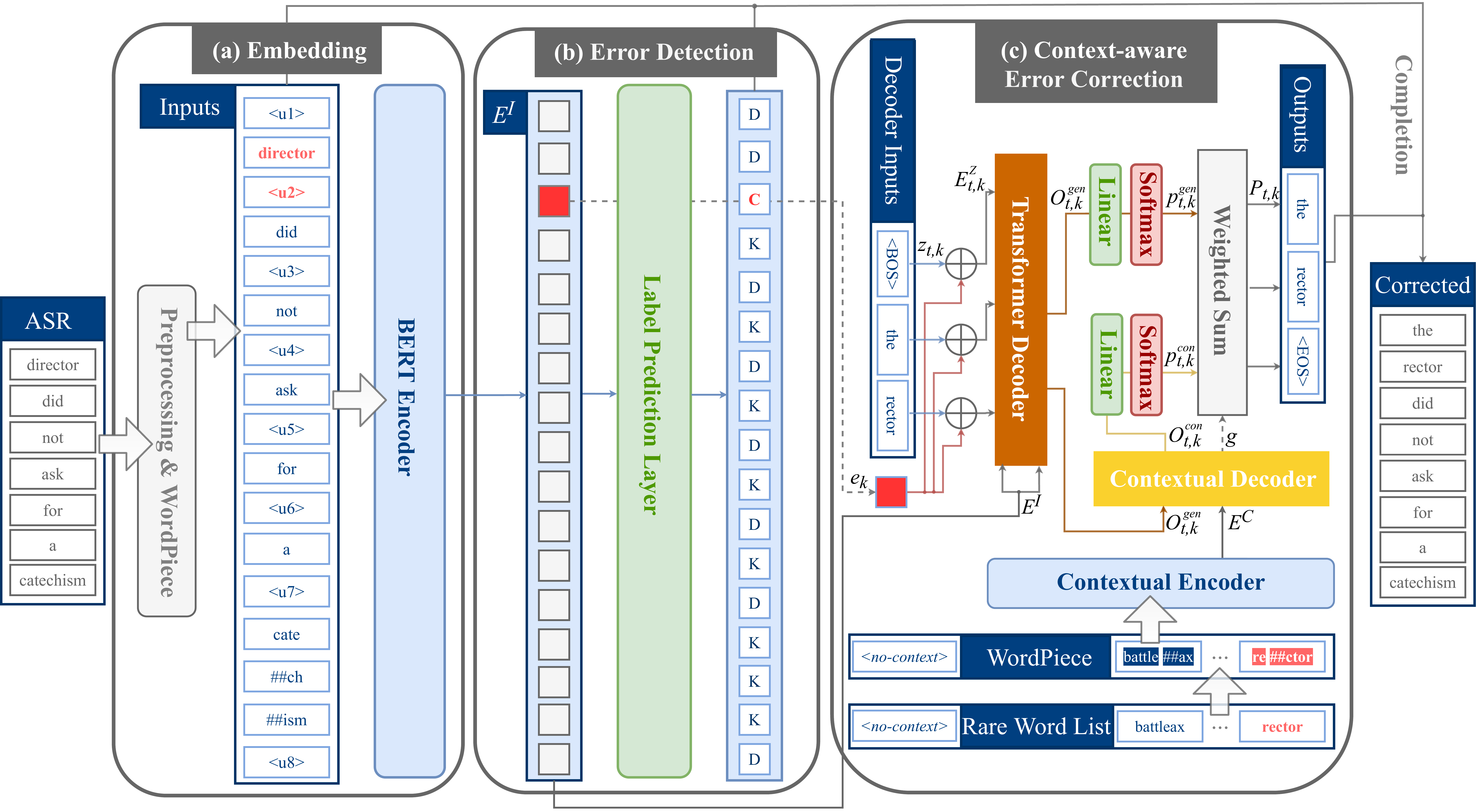}
  \vspace{-1mm}
  \caption{Overall architecture of the proposed ED-CEC model.}
  \vspace{-2mm}
  \label{fig:model}
\end{figure*}

\vspace{-1mm}
\section{Methodology}
\label{sec:method}
\vspace{-2mm}
In the previous method proposed by Yang et al. \cite{DBLP:journals/corr/abs-2208-04641}, a preprocessing step and an error detection module were introduced, which ensured inference speed while providing some error correction capability. However, it exhibited poor performance in correcting rare word errors. In this paper, we build upon their error detection module to preserve inference speed and introduce a novel context-aware error correction module to effectively handle rare word errors. This constitutes the main innovation of our study. We refer to the combined model as the ED-CEC model, depicted in Fig. \ref{fig:model}. In this section, we provide a detailed explanation of each module.


\vspace{-3mm}
\subsection{Embedding Module}
\label{section2.1}
\vspace{-1mm}

The contextual error correction problem can be represented as the mapping function $f(S,C) = T$, where the source $S = (s_1, s_2, \cdots , s_m)$ is the original ASR output, the context $C = (c_1, c_2, \cdots , c_l)$ is the rare word list containing $l$ contextual items, and the target $T = (t_1, t_2, \cdots , t_n)$ is the corrected transcript. All the tokens are applied to a predefined WordPiece vocabulary \cite{wu2016google}.
Similarly to \cite{DBLP:journals/corr/abs-2208-04641}, we first insert dummy tokens between every two consecutive words in $S$ to reduce the ambiguity of possible editing operations. We then align $S$ and $T$ by determining the longest common subsequence between them. By inserting these dummy tokens, we can generate an aligned representation of $S$ denoted as $I = (i_1, i_2, \cdots , i_{2m+1})$, where $m+1$ represents the number of inserted dummy tokens. Finally, the aligned tokens are labeled \textit{KEEP} (\textbf{K}), whereas the remaining tokens are labeled \textit{DELETE} (\textbf{D}) or \textit{CHANGE} (\textbf{C}). 
An example is illustrated in Fig. \ref{fig:model}.

The hidden representations  $E^I = (e_1, e_2, \cdots , e_{2m+1})$ of the model inputs $I$ are then obtained by using the pretrained language model BERT \cite{devlin2018bert} as an encoder:
\begin{equation}
\vspace{-2mm}
    E^I = {\rm BERT}({\rm TE}(I)+{\rm PE}(I)),
\label{eq3}
\end{equation}
where TE and PE denote the token embedding and position embedding, respectively. 

\vspace{-3mm}
\subsection{Error Detection Module}
\vspace{-1mm}

The label prediction layer is a straightforward fully connected network with three classes: \textbf{K}, \textbf{D}, and \textbf{C}. The impact of this module on the overall system size is minimal, making it a lightweight component. However, its contribution to improving the inference speed of the system is significant.
\begin{equation}
\label{eq2}
\vspace{-1mm}
  P(y_o|e_o) = {\rm Softmax}({\rm FC}(e_o)),
\end{equation}
where $e_o \in E^I$ and $y_o$, $o \in \{1, \cdots, 2m+1\}$ are the output of the BERT encoder and predicted labeling operations, respectively. $\rm{FC}$ is a fully connected layer.

\vspace{-3mm}
\subsection{Context-aware Error Correction Module}
\vspace{-1mm}
The context-aware error correction module plays a crucial role in correcting rare word errors and represents the main innovation of this study. Unlike conventional autoregressive decoders that start decoding from scratch, our decoder operates in parallel to the tokens predicted as \textbf{C}.
More specifically, once the \textbf{C} positions are identified, the decoder takes as input a sequence consisting of these tokens and their surrounding context. This input sequence is then fed into the transformer decoder to generate correction candidates for all the tokens requiring correction simultaneously.

For the $k^{th}$ change position, the decoding sequence can be represented as $Z_k = (z_{1,k}, z_{2,k}, \cdots , z_{T,k})$, where $T$ is the length of the decoding sequence, generated by the transformer decoder. We compute the decoder inputs at step $t$ as follows:
\begin{equation}
\vspace{-1mm}
  E_{t,k}^Z = {\rm FC}(({\rm TE}(z_{t,k})+{\rm PE}(z_{t,k})) \oplus e_k),
\end{equation}
where TE and PE are the same token embedding and position embedding as in Eq. (\ref{eq3}), respectively. $z_{1,k}$ is initialized by a special start token $<$\textit{BOS}$>$. $e_k$ is the output of the BERT encoder at the $k^{th}$ change position. ``$\oplus$" denotes a concatenate function. $\rm{FC}$ is the fully connected layer that maps the decoder inputs back to the same dimension as the embedding of $z_{t,k}$.
Then, a transformer decoder is applied to obtain the decoder layer output, where the query input $Q$ is the decoder input. Both the key input $K$ and the value input $V$ are the output of the BERT encoder of the model input $I$:
\vspace{-2mm}
\begin{equation}
Q = E_{t,k}^Z, K =E^I, V = E^I
\vspace{-5mm}
\end{equation}
\vspace{-1mm}
\begin{equation}
O_{t+1,k}^{gen} = {\rm Transformer_{Decoder}}(Q,K,V),
\end{equation}
where $O_{t+1,k}^{gen}$ is the decoder layer output. 
Finally, the generation output is calculated as:
\vspace{-1mm}

\begin{equation}
p_{t+1,k}^{gen} = {\rm Softmax}({\rm FC}(O_{t+1,k}^{gen})).
\end{equation}

To dynamically choose between selecting tokens from the rare word list and generating new tokens, we also introduce a novel contextual mechanism. We use the contextual mechanism comprising a contextual encoder and a contextual decoder, with the contextual decoder consisting of context attention and context-item attention. The following are the detailed descriptions:

\textbf{Contextual Encoder.} We store $l$ contextual items in our rare word list. The $j^{th}$ contextual item, denoted as $c_j = (c_j^1, \cdots, c_j^u)$, is represented by WordPiece tokenization mentioned above, where $u$ indicates the number of tokens in the respective contextual item. To optimize the model size and increase inference speed, we adopt parameter sharing between the BERT encoder and the contextual encoder. As a result, we utilize the identical BERT encoder to obtain the hidden representations of the contextual items:
%
\vspace{-2mm}
\begin{equation}
E^C={\rm BERT}({\rm TE}(C)+{\rm PE}(C)),
\vspace{-2mm}
\end{equation}
where $C$  is the rare word list mentioned in Section \ref{section2.1}. TE and PE are the token embedding and position embedding mentioned above. $E^C = (e_{1}^C, \cdots, e_{l}^C)$ is the output of the contextual encoder. 

\begin{figure}[htb]
\vspace{-4mm}
\begin{minipage}[b]{1.0\linewidth}
  \centering
  \centerline{\includegraphics[width=6cm]{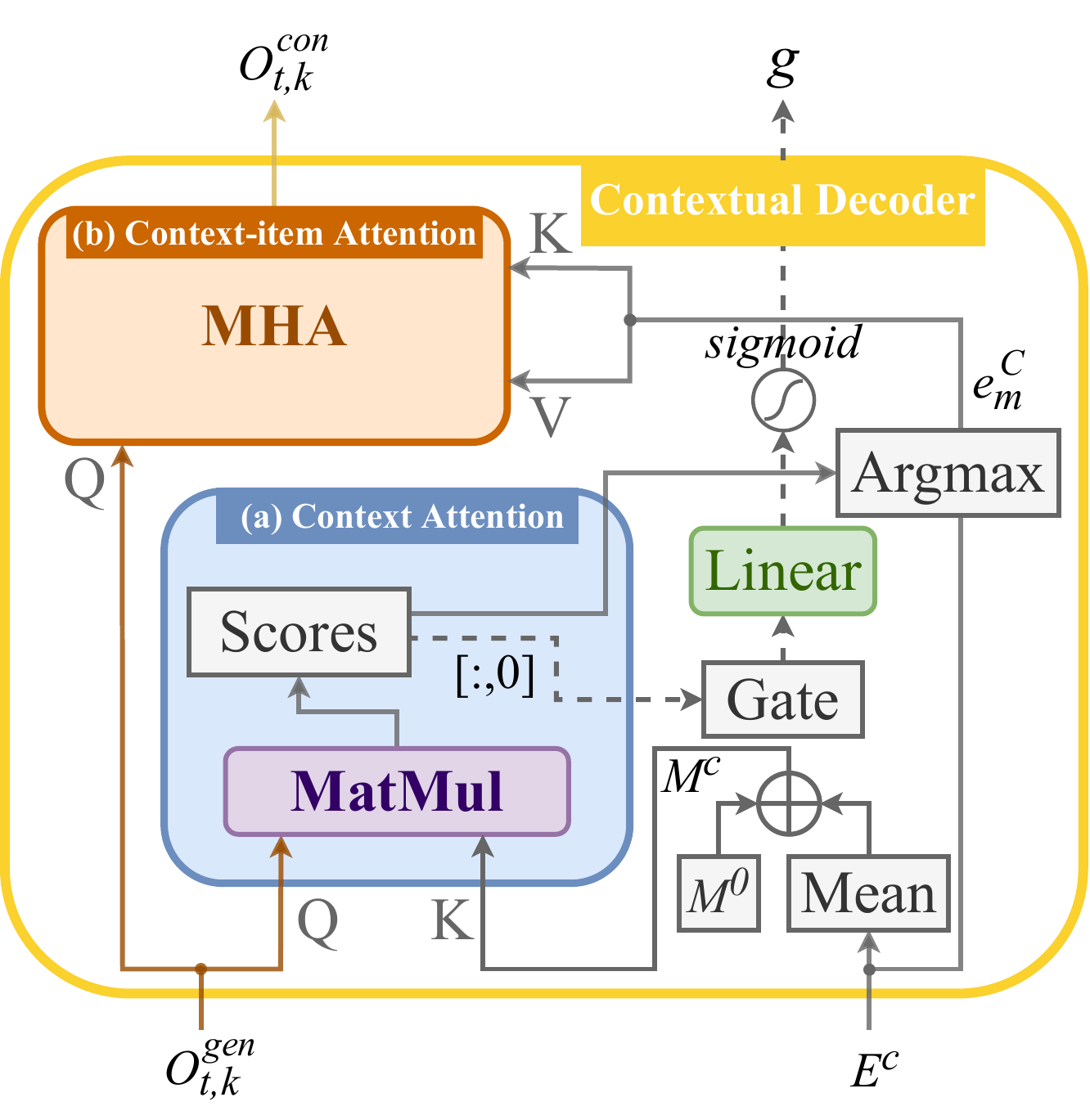}}
\end{minipage}
\vspace{-7mm}
\caption{Overall architecture of contextual decoder.}
\label{fig:res}
\vspace{-2mm}
\end{figure}

\textbf{Contextual Decoder.} The model diagram of the contextual decoder is shown in Fig. \ref{fig:res}. 
We compute the mean of each encoded contextual item and then expand these tokens with a learnable dummy token $<$\textit{no-context}$>$, which is utilized later to determine situations where there is no relevant knowledge stored in the rare word list:

\begin{equation}
M^C=M^0 \oplus {\rm mean}(E^C),
\end{equation}
%
where $M^0$ is the hidden representation of the learned dummy token $<$\textit{no-context}$>$  and $\rm{mean}$ is the mean operation on each $e_{j}^C \in E^C$, $j \in \{1, \cdots, l\}$. $M^C$ can be interpreted as summary tokens of contextual items.

At step $t$, the contextual decoder begins with a context attention layer that identifies the availability and location of a relevant contextual item in the rare word list by computing similarity scores, where the query input $Q_t$ and the key input $K_t$ are the output of the decoder layer and the summary tokens of contextual items at step $t$, respectively:
\vspace{-1mm}
\begin{equation}
Q_t = O_{t,k}^{gen}, K_t = M^C
\end{equation}
\vspace{-1mm}
\vspace{-7mm}

\begin{equation}
\vspace{-2mm}
\vspace{-1mm}
{\rm scores}_t = Q_tK_t^T
\vspace{-2mm}
\end{equation}

\begin{equation}
\vspace{-1mm}
gate_t = {\rm scores}_t[:,0],
\end{equation}
where ${\rm scores}_t[:,0]$, namely, $gate_t$ are the similarity scores corresponding to the $<$\textit{no-context}$>$ token $M^0$.
We define the index of the highest similarity score for the query $Q_t$ at step $t$ as $m = {\rm argmax}({\rm scores}_t)$. If $m$ is non-zero, indicating the presence of relevant contextual knowledge in the rare word list, we compute the contextual output using the context-item attention layer. This layer extracts the relevant information from a specific contextual item using a multihead attention (MHA) mechanism. In the MHA layer, the query input $Q_t$ is the output of the decoder layer, while the key input $K_t$ and the value input $V_t$ are both obtained from the contextual encoder of the $m^{th}$ contextual item at step $t$:
\vspace{-2mm}
\begin{equation}
Q_t = O_{t,k}^{gen}, K_t = e_m^C, V_t = e_m^C
\end{equation}
\vspace{-6mm}
\vspace{-3mm}
\begin{equation}
O_{t+1,k}^{con} = {\rm Softmax}( \frac{Q_tK_t^{T}}{\sqrt{d}}  )V_t
\end{equation}
\begin{equation}
p_{t,k}^{con} = {\rm Softmax}({\rm FC}(O_{t,k}^{con})),
\end{equation}
where the scaling factor $\sqrt{d}$ is for numerical stability. $O_{t,k}^{con}$ is the output of the context-item attention layer.

Then, the predicted word is acquired by a weighted sum between the generation output $p_{t,k}^{gen}$ and the contextual output $p_{t,k}^{con}$:
\vspace{-1mm}
\begin{equation}
g =  \sigma ({\rm FC}_{dim=1}(gate_t))
\end{equation}
\vspace{-5mm}
\begin{equation}
P_{t,k} = g \cdot p_{t,k}^{gen} + (1 - g) \cdot p_{t,k}^{con},
\end{equation}
where FC$_{dim=1}$ denotes one fully connected layer with one output neuron, $\sigma$ is the sigmoid function and $g$ is the gate to make a trade-off between the chosen token from the rare word list and the generated token.

\vspace{-3mm}
\subsection{Joint Training and Completion}
\vspace{-1mm}
The learning process is optimized through two objectives that correspond to error detection and context-aware error correction.
\vspace{-1mm}
\begin{equation}
{\rm Loss}_d = - \sum_{o}{\rm log}(P(y_o|i_o))
\vspace{-2mm}
\end{equation}
\begin{equation}
\begin{split}
{\rm Loss}_e = - (\sum_{k}\sum_{t}{\rm log}(P_{t,k}) + \\
\sum_{k}\sum_{t}{\rm log}(P({\rm label}_{t,k}|{\rm scores}_{t,k})),
\end{split}
\end{equation}
where the loss function ${\rm Loss}_d$ is the cross entropy loss for the detection network and the loss function ${\rm Loss}_e$ consists of two parts of the cross entropy loss for the context-aware correction network. Furthermore, ${\rm scores}_{t,k}$ are the score outputs of the context attention layer and ${\rm label}_{t,k}$ is the contextual label that contains the index of the corresponding contextual item. The two loss functions are linearly combined as the overall objective in the learning phase:
\vspace{-2mm}
%
\begin{equation}
{\rm Loss} = \gamma \cdot {\rm Loss}_d + {\rm Loss}_e,
\end{equation}
\vspace{-1mm}
where $\gamma$ is the hyperparameter for adjusting the weight between ${\rm Loss}_d$ and ${\rm Loss}_e$.

During the completion process, we convert the predicted operation labels and the generated words into a complete utterance. Specifically, as depicted in Fig. \ref{fig:model}, we preserve the tokens labeled \textbf{K} and remove those labeled \textbf{D} from the inputs. We then replace the tokens labeled \textbf{C} with the corresponding generated words.

\vspace{-1mm}
\section{Experimental EVALUATIONS}
\label{sec:experiment}
\vspace{-1mm}
\subsection{Experiment Settings}
\vspace{-1mm}
Our method was implemented using Python 3.7 and Pytorch 1.11.0. The model was trained and evaluated on a computer with Intel(R) Xeon(R) Gold 6248 CPU @ 2.50GHz, 32GB RAM, and one NVIDIA Tesla V100 GPU. 

Both the BERT encoder and the contextual encoder employed the same bert-base-uncased model \cite{devlin2018bert} for initialization. The vocabulary size for word tokenization was 30522. We set the hidden size as 768, the number of attention layers as 12, and the number of attention heads as 12. The transformer decoder adopted a single-layer transformer with a hidden size of 768. We used Adam \cite{kingma2014adam} as the optimizer with a batch size of 32 and set $\gamma$ to 3. The initial learning rate was 0.00005. All the hyperparameters were fine-tuned on the standard validation data.

\vspace{-2mm}
\subsection{Data}
\vspace{-1mm}
To assess the effectiveness and robustness of our proposed model, we employ five datasets that utilize various ASR engines. The statistics of the datasets are shown in Table \ref{tab:dataset}. 

\begin{itemize}[leftmargin=*]
\setlength{\itemsep}{0pt}
\setlength{\parsep}{0pt}
\setlength{\parskip}{0pt}
\vspace{-1mm}
\item The ATIS dataset \cite{sundararaman2021phoneme} includes 8 hours of audio recordings of people making flight reservations, along with their corresponding human transcripts. ASR transcripts were generated by a LAS ASR system \cite{chan2016listen}.

\item The SNIPS dataset \cite{9054689} is collected from the SNIPS voice assistant,  focusing on natural language understanding. The Kaldi\footnote{\url{https://github.com/kaldi-asr/kaldi}} ASR toolkit was used to obtain the corresponding transcripts.

\item The Librispeech dataset \cite{2015Librispeech} is a collection of 960 hours of audiobooks. The ESPNet \cite{watanabe2018espnet} ASR toolkit was utilized to obtain related transcripts. We used dev-clean and test-clean as the validation and test sets, respectively.

\item The MELD dataset \cite{poria2018meld} consists of more than 1400 dialogues and 13000 utterances extracted from the Friends TV series. We utilized Whisper \cite{radford2022robust} to obtain transcripts. 

\item The PRLVS dataset \cite{hernandez2021multimodal} comprises a complete semester course consisting of pattern recognition lecture videos accompanied by slides. The course consists of 43 videos, with a total duration of 11.4 hours. We employed SpeechBrain \cite{ravanelli2021speechbrain} to obtain the transcripts related to the videos.
\vspace{-3mm}

\end{itemize}

\begin{table}[h!]


  \centering
  \resizebox{\linewidth}{!}{\begin{tabular}{@{}|c|c|c|c|c|c|@{}}
    \hline
    Dataset&  ATIS &SNIPS  &Librispeech & MELD & PRLVS  \\
    \hline
     Train  & 3867& 13084 & 252691 & 9989& 3680  \\
      Valid& 967 &700&   2703 &1109 & 460    \\
   Test  &800 &700 &2620 &2610 & 460   \\
    \hline
    
  \end{tabular}}

  \caption{Numbers of utterances in different datasets.}
  \vspace{-6mm}
  \label{tab:dataset}
   
\end{table}

\subsection{Rare Word List Construction}
Owing to the lack of available rare word lists for the ATIS, SNIPS, Librispeech, and MELD datasets, we followed the approach proposed in \cite{le2021contextualized} to construct rare word lists for each dataset. Specifically,  a complete rare word list was initially compiled for the Librispeech dataset, consisting of 209.2K distinct words, by excluding the top 5,000 most common words from the Librispeech LM training corpus. Next, the rare word lists were constructed by identifying words from the reference of each utterance that were present in the complete rare word list. Additionally, a specified number of distractors (e.g., 1,000) were added to each rare word list, as determined by the experiment requirements. By utilizing this approach, we can effectively organize the rare word lists for each utterance\footnote{\url{https://github.com/facebookresearch/fbai-speech/tree/master/is21_deep_bias}}, containing words from the complete rare word list and supplementing them with distractors. Similar methods were employed to construct rare word lists for the remaining ATIS, SNIPS, and MELD datasets.

To demonstrate the feasibility of obtaining rare word lists in practice, we focused on the PRLVS dataset. The construction process involved collecting slides for each lecture and utilizing the Tesseract 4 OCR engine\footnote{\url{https://github.com/tesseract-ocr/tesseract}} for text extraction. Distinct word tokens were then extracted from the OCR output files. Among these tokens, only those belonging to the complete rare word list and appearing fewer than 15 times in the PRLVS train set were included in the lecture-specific rare word list. These rare word lists were subsequently applied for the context-aware correction of all utterances within the corresponding lectures \cite{sun2022tree}.

\vspace{-2mm}
\subsection{Baselines and Metrics}
\vspace{-1mm}

We evaluate the error correction performance of our proposed method, as well as four baselines:
\begin{itemize}[leftmargin=*]
\setlength{\itemsep}{0pt}
\setlength{\parsep}{0pt}
\setlength{\parskip}{0pt}
\vspace{-1mm}
\item Original denotes the original ASR output.
\item SC\_BART \cite{zhao2021bart} has demonstrated superior performance in ASR error correction tasks, achieving SOTA results.
\item distillBART \cite{shleifer2020pre} is a distilled version of the BART large model.
\item ConstDecoder$_{trans}$ \cite{DBLP:journals/corr/abs-2208-04641} is a constrained decoding method designed to improve the inference speed of ASR error correction while preserving a certain level of error correction performance.
\vspace{-1mm}
\end{itemize}

In addition, we use the following four evaluation metrics to assess the performance:
\begin{itemize}[leftmargin=*]
\setlength{\itemsep}{0pt}
\setlength{\parsep}{0pt}
\setlength{\parskip}{0pt}
\vspace{-1mm}
\item WER is the overall word error rate (WER) on all words.
\item WERR quantifies the WER reduction across all words.
\item U-WER calculates the unbiased WER on words not included in the rare word list.
\item B-WER computes the biased WER on words present in the rare word list.
\vspace{-1mm}
\end{itemize}

 In the case of insertion errors, if the inserted word is found in the rare word list, it will contribute to B-WER; otherwise, it will be considered for U-WER. The objective of contextualization is to improve B-WER while minimizing any significant degradation in the U-WER \cite{le2021contextualized}.

\begin{table}[htbp]
  
  \centering
\resizebox{\linewidth}{!}{
  \begin{tabular}{@{}|c|c|c|c|c|c|@{}}
    \hline
    \multirow{1}{*}{Method}  & ATIS & SNIPS & Librispeech  & MELD & PRLVS \\
    \hline
    SC\_BART  &90.30 &75.30  &152.93 & 25.70  & 103.62 \\
    distillBART  &45.55 &41.55  &73.29 & 11.99  & 57.60 \\
   ConstDecoder$_{trans}$   &25.61 &26.66    &17.10 &4.23  & 18.29\\
   \hline
   \hline
   ED-CEC (Proposed)      &32.59 &31.87    &25.48 &5.71  & 22.86\\
   vs SC\_BART & 2.8${\times}$ &2.4${\times}$    &6.0${\times}$   & 4.5${\times}$ &4.5${\times}$\\
   vs distillBART & 1.4${\times}$ &1.3${\times}$    &2.9${\times}$ &2.1${\times}$  & 2.51${\times}$\\
   vs ConstDecoder$_{trans}$ & 0.8${\times}$ &0.8${\times}$    &0.7${\times}$  &0.7${\times}$  & 0.8${\times}$\\
    \hline
  \end{tabular}}
  \caption{Average inference time in milliseconds (ms).}
  \vspace{-6mm}
  \label{tab:inference_speed}

\end{table}

\begin{table*}[htbp]

  \centering
  \resizebox{1.0\textwidth}{!}{
  \begin{tabular}{@{}|c|c|c|c|c|c|@{}}
    \hline
    \multirow{3}{*}{Method}  & ATIS & SNIPS & Librispeech   & MELD  & PRLVS \\
    \cline{2-6} 
     & WER/WERR & WER/WERR & WER/WERR & WER/WERR & WER/WERR\\
      & (U-WER/B-WER) & (U-WER/B-WER) & (U-WER/B-WER) & (U-WER/B-WER) & (U-WER/B-WER) \\
    \hline
    \multirow{2}{*}{Original} & 30.65/- & 45.73/-  &6.75/ - &31.08/ - &18.66/ -    \\
    & (20.58/ 87.78)& (34.20/99.64)&(3.13/30.26)&(25.31/ 74.62)  &(10.66/ 47.24 )  \\
    \multirow{2}{*}{SC\_BART} &21.47/29.95 &30.35/33.63   &5.78/14.37  &29.51/ 5.05  &15.14/ 18.86  \\
    & (14.63/49.25)& (21.86/70.25) &(3.80/18.45) &(24.25/67.73) &(10.43/27.67 )  \\
    \multirow{2}{*}{distillBART} &26.51/13.51 &33.28/27.23  &6.36/5.78  &30.76/1.03  &17.98/ 3.64  \\
    & (18.54/74.67)      & (24.08/76.43)      &(3.88/23.49)   &(25.06/73.89)    &(10.64/ 43.57)          \\
   \multirow{2}{*}{ConstDecoder$_{trans}$}   &21.74/29.07 &30.98/32.25 &5.89/12.74 &29.98/ 3.54&15.31/ 17.95 \\
   & (14.78/50.57)&(22.09/71.43)&(3.68/19.94)&(24.70/69.42)&(10.95/28.33) \\
   \hline
   \multirow{2}{*}{ED-CEC (Proposed)}      &\textbf{18.95}/\textbf{38.17} &\textbf{28.57}/\textbf{37.52}    &\textbf{5.08}/\textbf{24.74} &\textbf{26.23/15.60} &\textbf{13.17/ 29.42} \\
   & \textbf{(14.88/38.38)} & \textbf{(21.47/62.79)} &\textbf{(3.77/12.38)}&\textbf{(21.31/56.02)} &\textbf{(10.01/ 20.72 )}  \\
    \hline
  \end{tabular}
  }
  \caption{Measurements of error correction performance on five datasets.}
  \vspace{-4mm}
  \label{tab:wer_results}
\end{table*}

\vspace{-1mm}
\subsection{Results and Analysis}
\vspace{-1mm}
Tables \ref{tab:inference_speed} and \ref{tab:wer_results} provide evidence that our model achieves significant improvements in inference speed and WER results compared with the previous SOTA model on all five datasets when the rare word list size is set to 100. Compared with the original ASR output, our model achieves a marked WERR, ranging from 15.6$\%$ to 38.17$\%$. The average relative improvement in B-WER is 46.68$\%$. Furthermore, our model demonstrates considerable gains in inference speed, being 2.4 to 6.0 times higher than the previous SOTA model. This proves that our model achieves a good tradeoff between inference speed and WER. Additionally, the performance improvements across five different ASR systems prove the robustness of our model.

\begin{figure}[htbp]

\begin{minipage}[b]{1\linewidth}
\vspace{-3mm}
  \centering
  \centerline{\includegraphics[width=8cm]{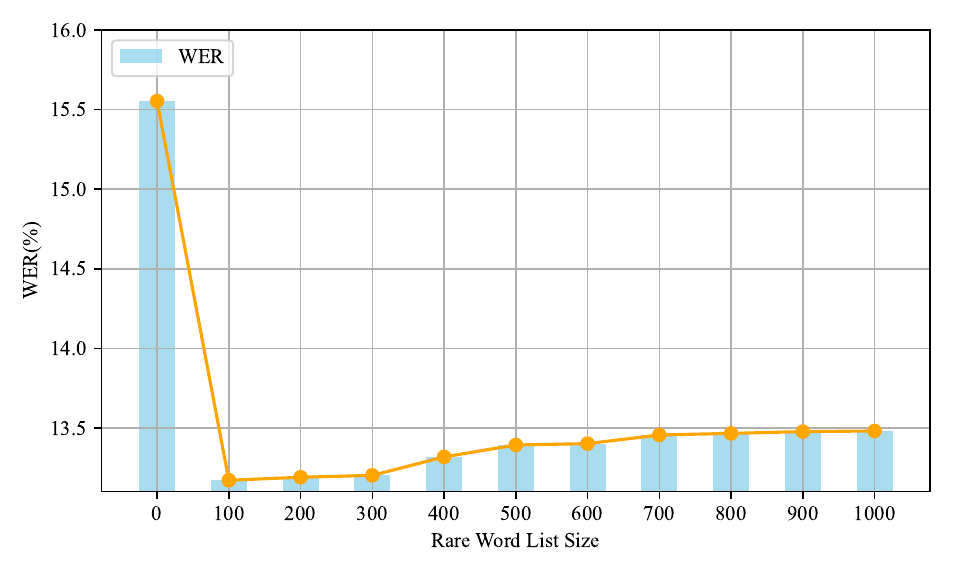}}
 \vspace{-5mm}
\end{minipage}
\caption{WER results of PRLVS with different rare word list sizes.}
\vspace{-1mm}
\label{fig:size}
\end{figure}

We also experimented on the PRLVS dataset with varying rare word list sizes, created by augmenting the rare word lists with distractors, ranging from 100 to 1000 contextual items.
The best WER results were observed with a rare word list size of 100, as shown in Fig. \ref{fig:size}. Increasing the rare word list size to 1000 showed a minor upward trend in WER, possibly due to false positives. Importantly, an empty rare word list resulted in a significant WER increase, highlighting the model's reliance on contextual items for rare word correction.
Furthermore, we conducted an “anti-context” experiment, employing a rare word list only containing 100 unrelated distractors. In this case, the WER was 15.45\%.
Thus, our approach yields optimal results when combined with a small number of relevant rare words that the model should prioritize.

\begin{figure}[h!]

\begin{minipage}[b]{1.0\linewidth}
  \centering
  \centerline{\includegraphics[width=9cm]{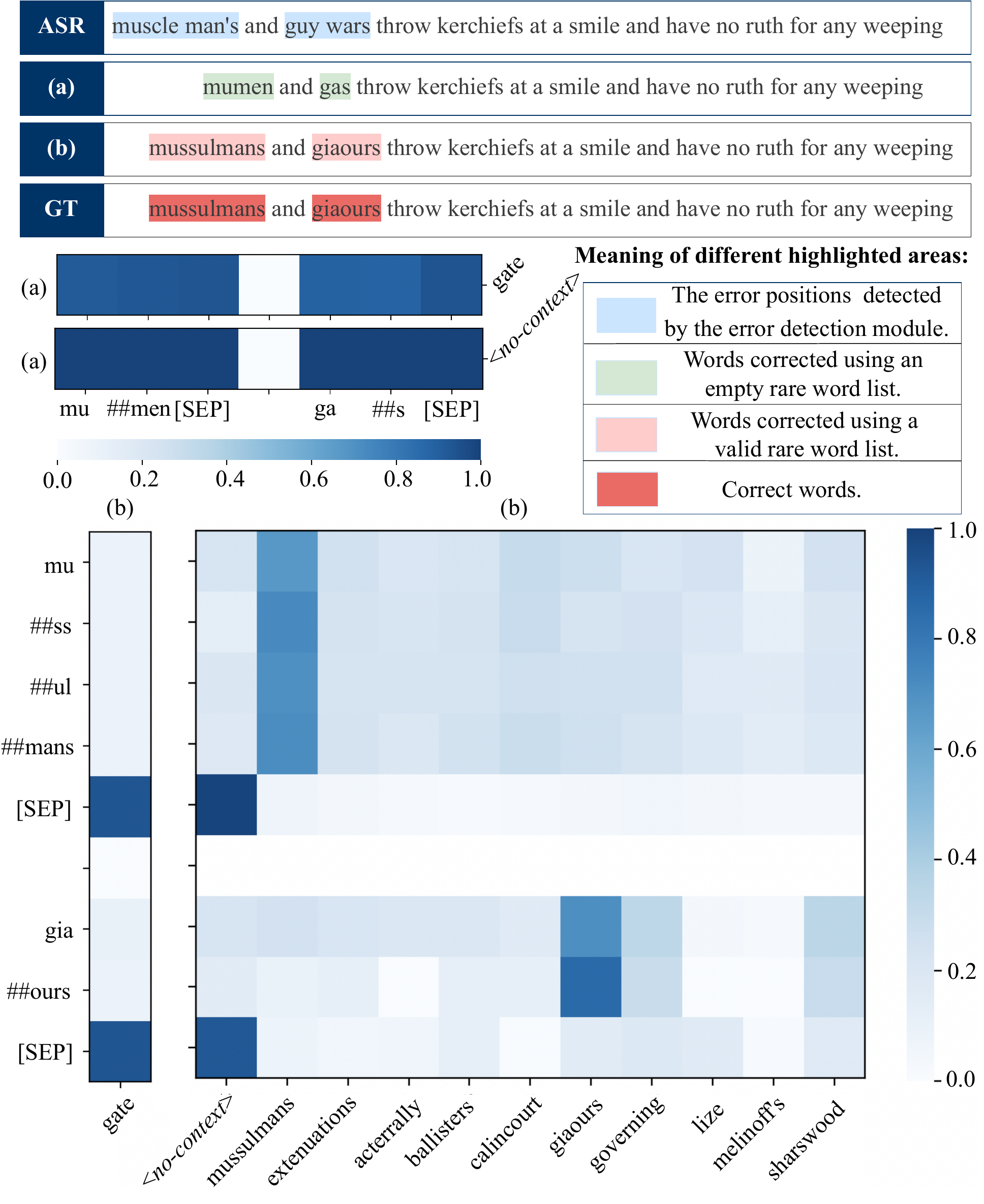}}
\end{minipage}
\vspace{-6mm}
\caption{Example for correcting rare words.}
\label{fig:heatmap}
%
\end{figure}

Fig. \ref{fig:heatmap} shows a specific example of correcting rare words ``mussulmans" and ``giaours". The \textbf{ASR} refers to the original ASR output. In \textbf{(a)}, decoding is performed with an empty rare word list, whereas in \textbf{(b)}, decoding is performed with a rare word list of size 10 containing the rare words ``mussulmans" and ``giaour". The \textbf{GT} denotes the ground truth transcript. When the rare word list is empty, the heat map of the gate (the weights between the original transformer decoder and the contextual decoder) indicates a stronger preference towards the output of the original transformer decoder. The generated words ``mumen" and ``gas" significantly deviate from the ground truth. However, when the rare word list contains the target rare words, the gate tends to favor the output of the contextual decoder, and the heat map of the rare word list shows that the model correctly selects the positions of the rare words.

\section{Conclusion And Future Work}
\label{sec:conclusion}
In this paper, we propose a fast and efficient contextual ASR error correction method that incorporates two main modules: an error detection module and a context-aware error correction module. This approach ensures both a high inference speed and the accurate correction of rare word errors in the ASR output. Experimental results on five datasets show the effectiveness and robustness of our model. In the future, we plan to extend our model by adding an additional phoneme encoder to recognize error patterns at the phoneme level, which will enable us to better bias rare words for correction.

\noindent \textbf{Acknowledgement.} This work was supported in part by JST CREST Grant Number JPMJCR22D1, Japan, and a project, JPNP20
-006, commissioned by NEDO.

\bibliographystyle{IEEEbib}
\bibliography{strings,refs}

\end{document}